\documentclass[10pt,twocolumn,letterpaper]{article}

\usepackage{iccv}
\usepackage{times}
\usepackage{epsfig}
\usepackage{graphicx}

\usepackage{ulem}
\usepackage{cuted}
\usepackage{amsfonts} 
\usepackage{amsmath}
\usepackage{array}
\usepackage{mathtools}
\usepackage{comment}
\usepackage{bbm}
\usepackage{bm}
\usepackage{multicol}
\usepackage{multirow}
\usepackage{threeparttable}
\DeclareMathAlphabet{\mathsfit}{\encodingdefault}{\sfdefault}{m}{sl}
\SetMathAlphabet{\mathsfit}{bold}{\encodingdefault}{\sfdefault}{bx}{n}
\newcommand{\tens}[1]{\bm{\mathsfit{#1}}}
\usepackage{color}
\usepackage{textcase}
\usepackage{bbding}
\usepackage{array}
\usepackage{booktabs}
\usepackage{enumitem}
\usepackage{arydshln}
\usepackage{makecell}
\usepackage{pifont}
\usepackage{soul}

\makeatletter
\renewcommand*\env@matrix[1][\arraystretch]{%
  \edef\arraystretch{#1}%
  \hskip -\arraycolsep
  \let\@ifnextchar\new@ifnextchar
  \array{*\c@MaxMatrixCols c}}
\makeatother

\iccvfinalcopy 

\def\iccvPaperID{} 

\ificcvfinal\fi

\begin{document}

\title{Learning Snippet-to-Motion Progression for Skeleton-based Human Motion Prediction}

\author{Xinshun Wang\\
Sun Yat-sen University\\
\and
Qiongjie Cui\\
Nanjing University of Science and Technology\\
\and
Chen Chen\\
University of Central Florida\\
\and
Shen Zhao\\
Sun Yat-sen University\\
\and
Mengyuan Liu\\
Peking University\\
}

\maketitle
\ificcvfinal\thispagestyle{empty}\fi

\begin{abstract}
    Existing Graph Convolutional Networks to achieve human motion prediction largely adopt a one-step scheme, which output the prediction straight from history input, failing to exploit human motion patterns.
    We observe that human motions have transitional patterns and can be split into snippets representative of each transition. Each snippet can be reconstructed from its starting and ending poses referred to as the transitional poses.
    We propose a snippet-to-motion multi-stage framework that breaks motion prediction into sub-tasks easier to accomplish.
    Each sub-task integrates three modules: transitional pose prediction, snippet reconstruction, and snippet-to-motion prediction.
    Specifically, we propose to first predict only the transitional poses. Then we use them to reconstruct the corresponding snippets, obtaining a close approximation to the true motion sequence.
    Finally we refine them to produce the final prediction output. 
    To implement the network, we propose a novel unified graph modeling, which allows for direct and effective feature propagation compared to existing approaches which rely on separate space-time modeling.
    Extensive experiments on Human 3.6M, CMU Mocap and 3DPW datasets verify the effectiveness of our method which achieves state-of-the-art performance.
\end{abstract}

\section{Introduction}
Human motion prediction has relevance to many applications in multimedia, human-robot interaction, autonomous driving \cite{ding2022towards, hou2014compressing, tang2023collaborative}, etc. It often involves multimedia techniques such as video processing \cite{liu2017high,zhang2018detecting,tu2023consistent,tu2022general,tu2023dtcm} and motion estimation \cite{liu2018recognizing} to build a more accurate representation of human movements.
Researchers have found early success with RNNs \cite{men2020quadruple,fragkiadaki2015recurrent,ghosh2017learning} and CNNs \cite{liu2020trajectorycnn,butepage2017deep,li2018convolutional}, which overlook structural information of skeletons.
The most engaging approaches are Graph Convolutional Networks (GCNs) \cite{kipf2016semi}, which can model articulated structures as graphs with nodes and edges.
While existing GCNs \cite{mao2019learning,cui2020learning,dang2021msr,li2020dynamic,sofianos2021space,li2021multiscale} in the field excel at capturing spatial relationships within individual frames or graphs, they often lack explicit mechanisms to effectively model the temporal evolution of human motion. Instead, these methods typically rely on aggregating information from different frames without explicitly considering the temporal context, such as by using temporal CNN \cite{cui2020learning} along the time axis. This limitation hinders their ability to capture the sequential patterns, subtle motion transitions, and nuanced temporal dependencies inherent in human movement.

    \begin{figure}[t]
    \centering
    \includegraphics[width=0.99\columnwidth]{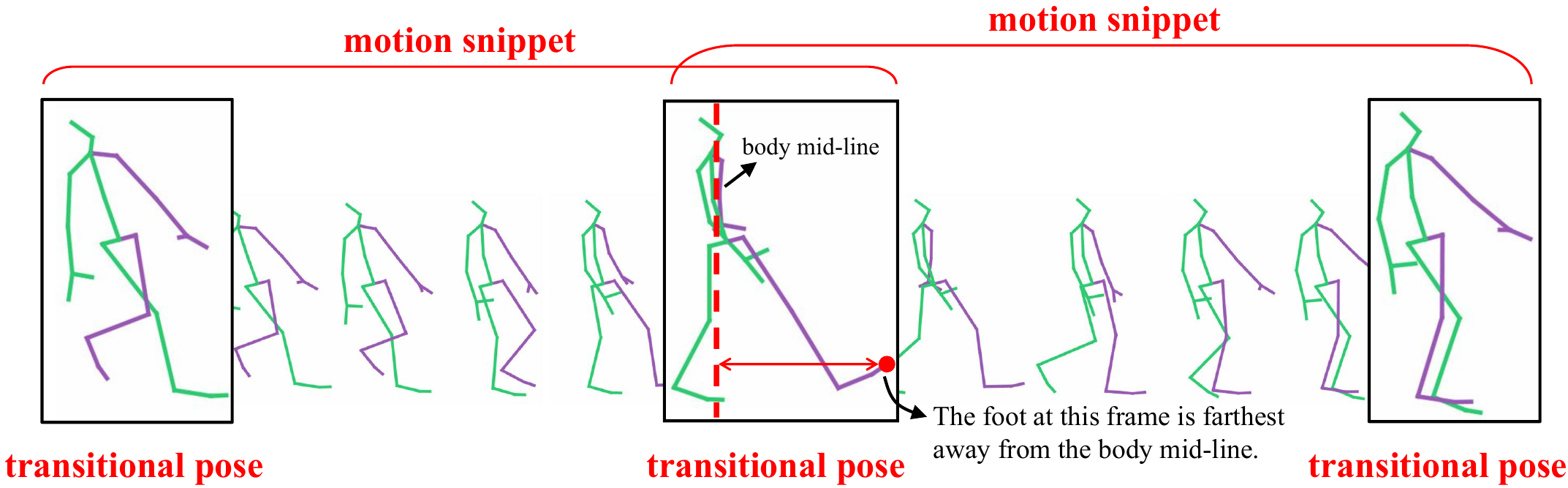}
    \caption{\small \small
    Illustration of ``transitional pose'' and ``motion snippet'' in a walking action case. We propose to first predict the transitional poses, then interpolate them to approximate the whole sequence, and finally refine them to produce the final prediction.}
    \label{Figure0_teaser_example}
    \end{figure}
    
To address this limitation, we propose a novel snippet-based approach incorporating a snippet-based motion representation and a snippet-to-motion prediction framework.
We are motivated by the observation that predicting a few poses is easier than predicting the whole sequence, and that human motion tends to exhibit multi-phase patterns. Existing methods do not exploit these observations.
As shown in Figure \ref{Figure0_teaser_example}, a motion sequence can be split by several transitional points into ``motion snippets'', each of a narrow time span, such that between any two subsequent points, it is possible to obtain a close approximation to the motion snippet only by interpolating between the starting and ending poses of this snippet.
Consider a simple walking motion for example. The transitional points would be when the hands (or feet) are farthest away from the mid-line of the body. The poses at these points are referred to as transitional poses.
The snippets are concatenated together to produce a realistic motion sequence that is fairly close to the true one.
To leverage this observation, we propose to first predict the future transitional points for each specific sample. Then we reconstruct from these transitional poses the corresponding motion snippets by using techniques such as linear interpolation. Finally, a snippet-to-motion prediction module assembles the snippets to obtain a realistic motion sequence, and then refine the sequence to produce the final predicted motion sequence.

The snippet construction scheme described above is sample-specific, which is denoted by a ``non-shared'' suffix. To save computational cost, we provide another scheme of constructing snippets, one that highlights the statistically more probable and common transitional patterns.
Specifically, after having obtained all the transitional points corresponding to each sample, the model will take their averages to be the transitional points for all samples.
For example, if sample 1 has three snippets split by 2 transitional poses at time $[t^{(1)}_1, t^{(1)}_2]$, and sample 2 at time $[t^{(2)}_1, t^{(2)}_2]$, then we use $[(t^{(1)}_1 + t^{(2)}_1)/2, (t^{(1)}_2 + t^{(2)}_2)/2]$ as the transitional points shared by all samples. The resulting model is denoted using a ``shared'' suffix.

We employ GCN to implement our network. Existing GCN-based approaches \cite{sofianos2021space,li2021multiscale,ma2022progressively} largely implement graph convolution separately in the spatial and temporal domains, which causes indirect information flow impairing their ability to effectively model long-range spatio-temporal dependencies.
To combat this issue, we propose a novel unified graph modeling to represent the whole motion sequence as a unified spatio-temporal graph. Such graph modeling allows for more direct and efficient cross-spacetime feature propagation over the graph representation.

Our contributions are three-fold:

\begingroup
\setlist[itemize,1]{leftmargin=0.37cm}
\begin{itemize}
    \item 
    We propose a framework integrating transitional pose prediction, snippet reconstruction and snippet-to-motion prediction modules, leveraging the observations that predicting a few poses is easier than predicting the whole sequence, and that human motion exhibits periodic and multi-phase patterns, which are unexplored in existing works.
    \item
    We propose a novel unified graph modeling, which enables the network to more efficiently and directly extract spatio-temporal features and learn long-range dependencies, facilitating snippet-to-motion progression at each stage.
    \item
    Extensive experiments on three challenging benchmark datasets, Human3.6M, CMU Mocap and 3DPW, consistently show that our proposed framework achieves state-of-the-art performances. The code can be found in the supplementary material.
\end{itemize}
\endgroup

\section{Related Work}

\subsection{GCNs in Human Motion Prediction}

Derived as a generalization of convolutions to non-Euclidean data \cite{bruna2013spectral,scarselli2008graph}, GCNs \cite{kipf2016semi,velivckovic2017graph, hamilton2017inductive, qi2017pointnet,bahdanau2014neural, liu2020disentangling, shi2021adasgn, li2019actional, shi2019skeleton, gao2019optimized} have become the favorite choice in this area. \cite{mao2019learning} leveraged GCNs into the field. 
Some early approaches \cite{cui2020learning,li2020dynamic,dang2021msr} implemented simple graph convolution depending on spatial adjacency to learn spatial-only dependencies with additional temporal CNNs (e.g., TCNs) \cite{lea2017temporal} employed to learn temporal-only dependencies, which cannot effectively model spatio-temporal relations. Some recent works \cite{sofianos2021space,li2021multiscale,ma2022progressively} implementing graph convolutions separately on the spatial and temporal graphs.
Researchers have fruitfully explored graph representation of graph-structured time series data \cite{yan2018spatial,guo2019attention,wu2019graph,yu2017spatio,li2019spatio,aoun2014graph,herzig2019spatio}.
Regarding motion skeleton sequence,
LTD \cite{mao2019learning} proposes to represent the motion sequence as an implicit fully-connected graph in trajectory space.
LDR \cite{cui2020learning} introduces a pose graph with a predefined topology based on natural adjacency.
There is a growing preference for multi-scale graphs \cite{li2020dynamic,dang2021msr,li2021multiscale,zhou2021learning}.
DMGNN \cite{li2020dynamic} uses dynamic multi-scale graphs to represent body segments at different scales.
MSR-GCN \cite{dang2021msr} employ MLPs to reduce or increase the number of joints to construct the multi-scale graph.
What differentiates our work from existing approaches is that we build a unified graph from a global perspective instead of trajectories or poses, which facilitates the proposed framework in better extracting features and capturing spatio-temporal dependencies.

\subsection{Frameworks for Human Motion Prediction}
Alternatively, some researchers have turned their eyes to better prediction strategies and designing prediction frameworks more suitable for motion prediction \cite{mao2020history, dang2021msr, ma2022progressively, Kiciroglu2020LongTM} instead of focusing on proposing new graph convolutions. This was especially the case after PGBIG \cite{ma2022progressively}, which introduced a multi-stage prediction framework, where each stage gradually provides better initial guesses for the next stage. \cite{mao2020history} employ the attention mechanism in conjunction with GCNs.
\cite{dang2021msr} propose a multi-stage prediction framework, where the skeleton is gradually downsampled and then upsampled using MLPs after several stages to obtain a series of fine-to-coarse graphs. Then the model learns to predict the future motion sequence accordingly.
\cite{Kiciroglu2020LongTM} design a RNN-based network to predict key poses for every motion sample and approximate intermediate ones by interpolating the key poses. Our approach differs from their method in that we employ GCNs to implement the prediction model, and that instead of predicting classification labels and duration between key poses, we propose to directly predict future transitional poses which better exploits transitional patterns of human motions.

\section{The Proposed Method}

\subsection{Preliminaries}

Suppose that a motion sequence consists of $T$ consecutive frames, where each pose has $J$ joints. Then the motion sequence is denoted by $\tens{X}_{-H:T-1} = [\mathbf{X}_{-H}, \mathbf{X}_{-H+1}, \cdots, \mathbf{X}_{T-1}] \in \mathbb{R}^{(T+H)\times J \times D}$, where $\mathbf{X}_t \in \mathbb{R}^{J\times D}$ represents the pose at $t$-th frame, and each joint has $D$ features. Given the observed $H$-frame motion sequence $\tens{X}_{-H:-1}$, the human motion prediction task is to generate future $T$-frame motion sequence $\tens{X}_{0:T-1}$.
The graph convolution is defined as:
\begin{equation}
    \mathbf{H}^{(k)} = \sigma(\mathbf{A} \mathbf{X}^{(k-1)} \mathbf{W}^{(k)}),
\end{equation}
where $\mathbf{A}$ is the adjacency matrix, $\mathbf{X}^{(k)}$ is node features at $k$-th layer, $\mathbf{W}$ is the weight matrix, and $\sigma$ denotes an activation such as ReLU.

\subsection{Snippet-to-Motion Prediction Framework} \label{snippet prediction construction}

\textbf{Transitional Pose Prediction.}
First let us introduce the transitional pose prediction pipeline.
For each sample, the motion sequence is split into $S$ motion snippets:
$\tens{X}_{0:T_1}, \tens{X}_{T_1:T_2}, \cdots, \tens{X}_{T_{S-1}:T-1}.$
Note that there exists an one-frame overlap between every two consecutive snippets, which is a transitional pose. So the transitional poses are $\mathbf{X}_{T_1},\mathbf{X}_{T_2},\cdots, \mathbf{X}_{T_{S-1}}$.
The duration of each snippet is variable for different samples to ensure flexibility. We now define a reconstruction error as:

\begin{equation}\label{reconstruction error}
    \mathcal{R} (i) = \frac{1}{J} \sum_{j=1}^J \big\lVert \big[\hat{\mathbf{X}}_i\big]_{j,:} - \big[\mathbf{X}_i\big]_{j,:} \big\rVert_2,
\end{equation}
where matrix $\mathbf{X}_i\in\mathbb{R}^{J\times D}$ denotes the pose at $i$-th frame with $J$ joints each represented by $D$ features.

    \begin{figure*}[!ht]
    \centering
    \includegraphics[width=0.99\textwidth]{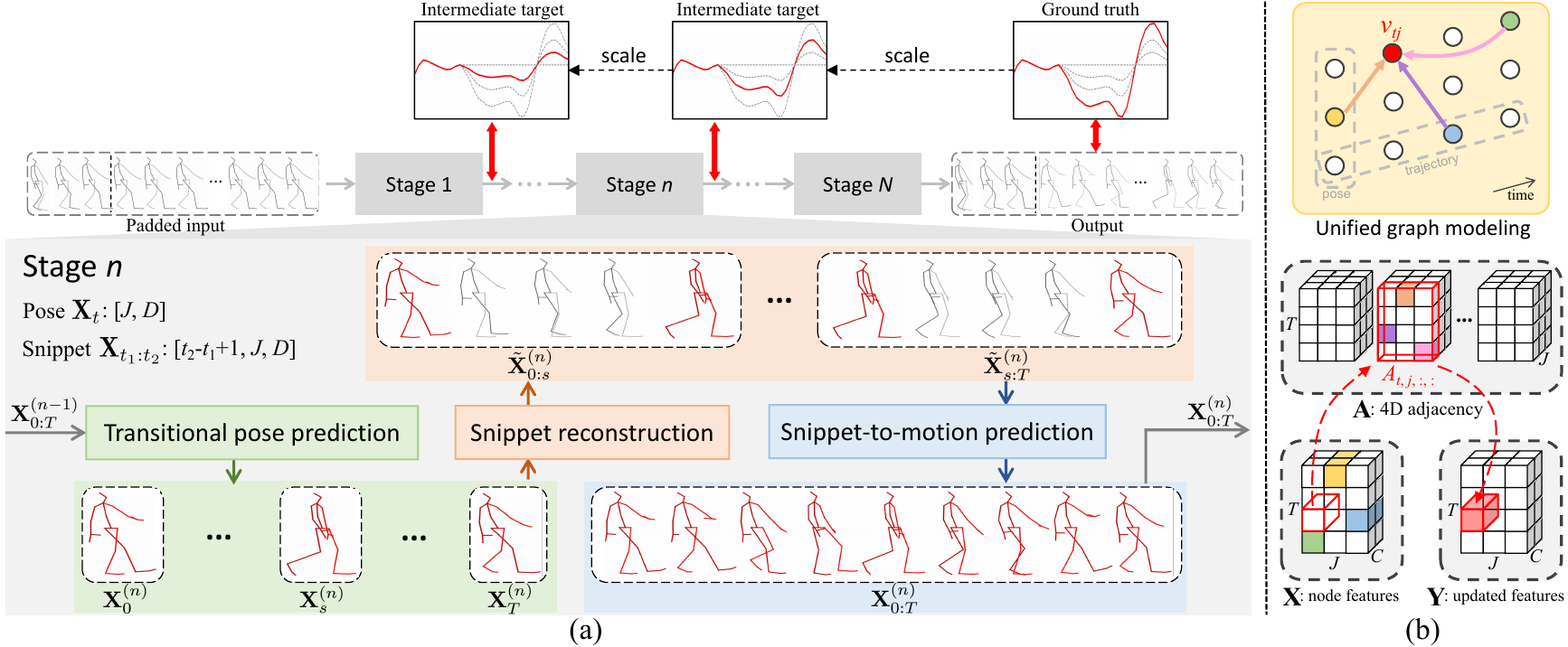} 
    \caption{\small Overview of the proposed snippet-to-motion prediction framework with intermediate supervision. (a) The framework adopts a multi-stage design with each stage accomplishing three tasks in sequential order. The first one learns to predict the transitional poses. The second one reconstructs each snippet from the corresponding predicted transitional poses. The last one concatenates all the snippets together and refine them to obtain the whole predicted sequence. Intermediate supervision is employed to guide the learning of transitional pose prediction and snippet-to-motion prediction, both implemented using graph convolutional networks with specific modifications made for predicting transitional poses. (b) The proposed unified graph modeling based on which we implement the graph network.}
    \label{model network framework architecture}
    \end{figure*}

Formally, for each sample we obtain its corresponding $S$ motion snippets by the following procedure:
\begin{enumerate}
    \item
    Perform linear interpolation between $\mathbf{X}_{-H}$ and $\mathbf{X}_{-1}$ to obtain an initial snippet, denoted by $\hat{\tens{X}}_{-H:-1}$, of the same duration as the motion sequence.
    \item
    To locate the transitional points, find frame index $i\in\{ -H,-H+1,\cdots,-1 \}$ at which the reconstruction error (as defined in Eq. \ref{reconstruction error}) between the pose $\hat{\mathbf{X}}_i$ and the corresponding ground truth $\mathbf{X}_i$ is maximized:
    \begin{equation}
        \text{argmax}_{i\in\{ -H,-H+1,\cdots,-1 \}}  \mathcal{R} (i)
    \end{equation}
    \item
    Obtain the initial transitional pose $\mathbf{X}_i$.
    \item
    Linearly interpolate between $\mathbf{X}_{-H}$ and $\mathbf{X}_{i}$, $\mathbf{X}_{i}$ and $\mathbf{X}_{-1}$, to obtain the new motion snippets $\hat{\tens{X}}_{-H:i}$ and $\hat{\tens{X}}_{i:-1}$.
    \item
    Find frame indexes to obtain the transitional poses over $\tens{X}_{-H:i}$ and $\tens{X}_{i:-1}$ accordingly.
    \item
    Iterate the steps above to obtain $S$ motion snippets.
\end{enumerate}

The number of motion snippets $S$ depends on the number of iterations and is a hyper-parameter that can be adjusted for model performance/complexity trade-off.
Having obtained the history motion snippets, we estimate the transitional points in the unknown future using a linear projection to obtain $T_1, T_2, \cdots, T_{S-1}$.
The snippet construction discussed above adopts a dynamic way of modeling human motion in the sense that it captures the unique transitional pattern defining each individual sample.

Finally, the transitional pose prediction task is to predict the transitional poses $\mathbf{X}_{0}, \mathbf{X}_{T_1}, \cdots, \mathbf{X}_{T_{S-1}}, \mathbf{X}_{T-1}$ using a graph convolutional network.
As discussed earlier, existing GCN-based models \cite{cui2020learning,dang2021msr,li2020dynamic,sofianos2021space,li2021multiscale} in human motion prediction share the same limitation in that they require the input and output of each graph convolutional layer to be of the same shape. As a result, they cannot be directly applied to the transitional pose prediction, because what we need is only several transitional poses, not the whole sequence. To this end, we propose a novel modification for GCN to make it applicable to our transitional pose prediction task.

Let us now focus on the last graph convolutional layer of the proposed network, which projects the hidden features of some $d$ dimensions corresponding to each node from the feature space of $\mathbb{R}^{T\times J\times d}$ into 3D space $\mathbb{R}^{T \times J\times D}$, producing the final prediction results. Assuming the last layer of the network is denoted by $K$, the graph convolution is written in matrix operations as:
\begin{equation}
    \mathbf{H}^{(K)} = \mathbf{A}^{(K)} \mathbf{H}^{(K-1)} \mathbf{W}^{(K)},
\end{equation}
where $\mathbf{A}\in\mathbb{R}^{(TJ)\times (TJ)}$ and $\mathbf{W}\in\mathbb{R}^{d\times D}$ are the adjacency matrix and projection matrix respectively. $\mathbf{H}\in\mathbb{R}^{(TJ)\times d}$ is the matrix of hidden features reshaped from $T\times J\times d$ into $TJ\times d$.
The pose-wise prediction result is described by the following equation:
\begin{equation}\label{pose-wise graph convolution}
    \big[\mathbf{H}^{(K)}\big]_{(tJ),:} = \big[\mathbf{A}^{(K)}\big]_{(tJ),:} \mathbf{H}^{(K-1)} \mathbf{W}^{(K)},
\end{equation}
where $t$ denotes the pose at $t$-th frame, and $[\cdot]_{(tJ),:}$ denotes the $(tJ)$-th row vector of the matrix.
In conclusion, the transitional poses can be obtained by assigning different values to $t$ in Eq. (\ref{pose-wise graph convolution}):
\begin{equation}
    \begin{split}
        & \mathbf{X}_t = \big[\mathbf{H}^{(K)}\big]_{(tJ),:},
    \end{split}
\end{equation}
This formalism allows the network to output only the poses we want, instead of all of them. 
As a result, the network is no longer subject to the limitation that the input and the output have to be equal in size, which brings more efficiency and more flexibility.
Therefore, We can specify the poses at certain frames that we want to predict, facilitating transitional pose prediction.

\noindent\textbf{Snippet Reconstruction.}
In the preceding step, we have obtained the predicted transitional poses representative of the whole sequence. Next the model reconstructs each motion snippet from the corresponding predicted transitional poses.
Formally, for the $s$-th snippet the reconstruction step can be written as:
\begin{equation}
    \hat{\tens{X}}_{T_{s-1}:T_{s}} = \textsc{reconstruct} (\hat{\mathbf{X}}_{T_{s-1}}, \hat{\mathbf{X}}_{T_{s}})
\end{equation}
The \textsc{reconstruct} function takes the starting and ending poses of each snippet as input, and reconstructs the motion snippets. Specifically, the $s$-th reconstructed motion snippet for each sample is obtained by implementing the \textsc{reconstruct} function as follows:
\begin{equation}
    \hat{\mathbf{X}}_t = \frac{\hat{\mathbf{X}}_{T_{s-1}} (T_s-t) + \hat{\mathbf{X}}_{T_s} (t-T_{s-1}) }{T_s-T_{s-1}},
\end{equation}
where $t = T_{s-1}, T_{s-1}+1, \cdots, T_s$.
When put altogether, the reconstructed snippets that we have obtained so far throughout the earlier steps represent a fairly close approximation to the truth, with only some refinements needed to fill the gap in between.

    \begin{table*}[t]
    \centering
    \resizebox{\textwidth}{!}{
    \begin{tabular}{p{2.6cm}|cccc|cccc|cccc|cccc}\hline
    scenarios  &\multicolumn{4}{c|}{Walking} &\multicolumn{4}{c|}{Eating} &\multicolumn{4}{c|}{Smoking} &\multicolumn{4}{c}{Discussion} \\ \hline
    milliseconds &80 &400 &560 &1000 &80 &400 &560 &1000 &80 &400 &560 &1000 &80 &400 &560 &1000\\ \hline
    LTD \cite{mao2019learning}	&9.25 	&33.75 	&43.83 	&48.97 	&8.80 	&46.74 	&57.91 	&70.75 	&7.47 	&30.16 	&33.81 	&63.42 	&10.68 	&44.80 	&75.00 	&115.09 	\\
    PGBIG \cite{ma2022progressively}	&8.39 	&31.22 	&39.04 	&45.76 	&8.66 	&50.38 	&57.56 	&69.99 	&6.02 	&30.20 	&33.27 	&59.05 	&8.09 	&44.52 	&72.55 	&98.85 	\\ 
    siMLPe \cite{guo2023back}	&7.85 	&34.43 	&41.44 	&45.60 	&8.52 	&47.08 	&62.06 	&74.85 	&6.49 	&32.26 	&38.44 	&63.12 	&8.41 	&53.18 	&76.50 	&100.63 	\\ \hline
    Ours (non-shared)	&7.80	&33.52	&42.88	&46.89	&8.47	&46.66	&59.36	&71.22	&6.39	&30.45	&33.90	&\textbf{\textcolor{blue}{56.03}}	&8.09	&47.01	&71.21	&103.88	\\
    Ours (shared)	&\textbf{\textcolor{blue}{7.36}}	&33.70	&41.11	&\textbf{\textcolor{blue}{44.95}}	&\textbf{\textcolor{blue}{8.10}}	&48.21	&\textbf{\textcolor{blue}{57.22}}	&\textbf{\textcolor{blue}{69.13}}	&\textbf{\textcolor{blue}{6.01}}	&29.81	&\textbf{\textcolor{blue}{32.82}}	&58.28	&\textbf{\textcolor{blue}{8.02}}	&50.99	&\textbf{\textcolor{blue}{71.07}}	&99.92	\\\hline

    \end{tabular}}
    \resizebox{\textwidth}{!}{
    \begin{tabular}{p{2.6cm}|cccc|cccc|cccc|cccc}\hline
    scenarios &\multicolumn{4}{c|}{Directions} &\multicolumn{4}{c|}{Greeting} &\multicolumn{4}{c|}{Phoning} &\multicolumn{4}{c}{Posing} \\ \hline
    milliseconds  &80 &400 &560 &1000 &80 &400 &560 &1000 &80 &400 &560 &1000 &80 &400 &560 &1000\\ \hline
    LTD \cite{mao2019learning}	&13.41 	&62.66 	&79.51 	&103.34 	&14.13 	&83.64 	&98.58 	&93.17 	&11.96 	&44.60 	&64.88 	&112.10 	&9.67 	&84.70 	&109.88 	&208.63 	\\
    PGBIG \cite{ma2022progressively}	&10.38 	&62.36 	&89.28 	&110.73 	&13.17 	&83.17 	&92.95 	&88.76 	&10.52 	&44.85 	&61.84 	&115.15 	&7.05 	&78.77 	&104.37 	&205.63 	\\ 
    siMLPe \cite{guo2023back}	&10.04 	&64.65 	&87.64 	&105.43 	&13.10 	&90.07 	&99.23 	&96.32 	&10.40 	&44.87 	&62.86 	&112.29 	&7.21 	&82.75 	&108.51 	&203.78 	\\ \hline
    Ours (non-shared)	&10.31	&62.46	&78.11 	&105.43 	&13.33	&81.89	&96.73 	&91.88 	&11.45	&45.38	&68.24 	&110.90 	&7.14	&81.21	&122.27 	&205.45 	\\
    Ours (shared)	&\textbf{\textcolor{blue}{9.66}}	&63.03	&\textbf{\textcolor{blue}{77.87}}	&108.05	&\textbf{\textcolor{blue}{12.15}}	&\textbf{\textcolor{blue}{81.19}}	&\textbf{\textcolor{blue}{91.73}}	&\textbf{\textcolor{blue}{87.96}}	&10.88	&\textbf{\textcolor{blue}{41.66}}	&64.53	&\textbf{\textcolor{blue}{108.96}}	&\textbf{\textcolor{blue}{7.02}}	&\textbf{\textcolor{blue}{77.93}}	&108.23	&\textbf{\textcolor{blue}{203.00}}	\\\hline

    \end{tabular}}
    \resizebox{\textwidth}{!}{
    \begin{tabular}{p{2.6cm}|cccc|cccc|cccc|cccc}\hline
    scenarios   &\multicolumn{4}{c|}{Purchases} &\multicolumn{4}{c|}{Sitting} &\multicolumn{4}{c|}{Sitting Down} &\multicolumn{4}{c}{Taking Photo}\\ \hline
    milliseconds &80 &400 &560 &1000 &80 &400 &560 &1000 &80 &400 &560 &1000 &80 &400 &560 &1000\\ \hline
    LTD \cite{mao2019learning}	&19.80 	&71.09 	&92.09 	&122.40 	&10.98 	&65.36 	&84.57 	&115.64 	&11.16 	&64.13 	&83.16 	&130.93 	&7.40 	&53.42 	&75.18 	&91.81 	\\
    PGBIG \cite{ma2022progressively}	&17.61 	&70.48 	&94.17 	&125.72 	&9.62 	&58.84 	&78.28 	&111.92 	&9.90 	&63.24 	&81.88 	&125.42 	&5.64 	&53.89 	&74.66 	&93.07 	\\
    siMLPe \cite{guo2023back}	&18.18 	&75.32 	&93.32 	&122.08 	&9.65 	&66.45 	&85.02 	&116.45 	&10.94 	&63.30 	&80.13 	&124.09 	&5.38 	&52.16 	&74.90 	&93.96  	\\   \hline
    Ours (non-shared)	&18.38	&74.67	&93.11	&122.87	&\textbf{\textcolor{blue}{9.48}}	&55.92	&79.16	&\textbf{\textcolor{blue}{109.88}}	&10.86	&60.74	&82.89	&128.97	&5.72	&53.12	&78.79	&93.41	\\
    Ours (shared)	&\textbf{\textcolor{blue}{17.55}}	&78.51	&92.45	&\textbf{\textcolor{blue}{121.19}}	&9.59	&\textbf{\textcolor{blue}{55.52}}	&\textbf{\textcolor{blue}{78.01}}	&115.48	&10.29	&\textbf{\textcolor{blue}{60.53}}	&80.79	&\textbf{\textcolor{blue}{124.08}}	&5.67	&\textbf{\textcolor{blue}{49.78}}	&\textbf{\textcolor{blue}{74.43}}	&99.69	\\\hline
    \end{tabular}}
    \resizebox{\textwidth}{!}{
    \begin{tabular}{p{2.6cm}|cccc|cccc|cccc|cccc}\hline
    scenarios   &\multicolumn{4}{c|}{Waiting} &\multicolumn{4}{c|}{Walking Dog} &\multicolumn{4}{c|}{Walking Together} &\multicolumn{4}{c}{Average} \\ \hline
    milliseconds &80 &400 &560 &1000 &80 &400 &560 &1000 &80 &400 &560 &1000 &80 &400 &560 &1000\\ \hline
    LTD \cite{mao2019learning}	&9.58 	&72.02 	&105.35 	&168.80 	&32.05 	&119.24 	&138.37 	&163.56 	&9.65 	&45.40 	&69.15 	&81.98 	&12.40 	&61.45 	&80.75 	&112.71 	\\
    PGBIG \cite{ma2022progressively}	&8.08 	&69.80 	&94.15 	&166.72 	&30.03 	&121.30 	&136.75 	&180.93 	&8.06 	&42.84 	&56.75 	&81.54 	&10.75 	&60.39 	&77.83 	&111.95 	\\
    siMLPe \cite{guo2023back}	&8.48 	&73.28 	&96.79 	&171.79 	&24.95 	&124.82 	&133.23 	&170.52 	&7.58 	&39.90 	&52.36 	&73.31 	&10.48 	&62.97 	&79.49 	&111.61 	\\     \hline
    Ours (non-shared)	&8.06	&68.25	&95.14	&\textbf{\textcolor{blue}{164.55}}	&24.10	&126.39	&134.45	&170.14	&7.78	&44.97	&58.99	&81.00	&10.49	&60.84	&79.68	&110.83	\\
    Ours (shared)	&\textbf{\textcolor{blue}{7.71}}	&\textbf{\textcolor{blue}{67.45}}	&\textbf{\textcolor{blue}{93.30}}	&165.16	&\textbf{\textcolor{blue}{22.77}}	&121.67	&\textbf{\textcolor{blue}{131.72}}	&\textbf{\textcolor{blue}{162.47}}	&\textbf{\textcolor{blue}{7.30}}	&\textbf{\textcolor{blue}{38.84}}	&58.16	&79.34	&\textbf{\textcolor{blue}{10.00}}	&\textbf{\textcolor{blue}{59.65}}	&\textbf{\textcolor{blue}{76.90}}	&\textbf{\textcolor{blue}{109.84}}	\\\hline
    \end{tabular}}
    \caption{\small Average MPJPEs (Mean Per Joint Position Errors) for long-term prediction on H3.6M. Note that the results provided in the original papers differ in testing strategies. Therefore we re-trained the models and evaluate them following the same paradigm as \cite{mao2019learning,li2020dynamic,martinez2017human}. ``shared'' and ``non-shared'' indicate whether or not the transitional points are shared among all samples in transitional pose prediction step.}
    \label{h36m_pred}
    \end{table*}

\noindent\textbf{Snippet-to-Motion Prediction.}
To implement our framework, we propose a GCN leveraging a novel unified graph modeling.
The general idea is that we do not construct the graph to model poses or trajectories, as is typical of existing GCN-based methods, but to model the whole motion sequence from a global perspective.

For each motion sequence sample with $T$ frames and $J$ joints, we construct a graph $G=(\mathcal{V}, \mathcal{E})$ to represent it, where the node set $\mathcal{V}$ has $T\times J$ nodes with each specifying a joint at a certain frame, and $\mathcal{E}$ denotes the set of edges connecting the nodes. Specifically, the graph is constructed under the following conditions: (1) There exists an edge connecting two nodes at the same frame if they are connected by a bone in the skeleton; (2) There exists an edge connecting two nodes at different frames if the corresponding joints are connected by a bone in the skeleton.
The graph can be fully specified by an adjacency matrix $\mathbf{A}\in\mathbb{R}^{TJ\times TJ}$.
And we implement the graph convolutional network depending on $\mathbf{A}$, with the last layer specifically reinvented to meet our needs as discussed earlier in Subsection \ref{snippet prediction construction}.
In practice the adjacency matrix is set as trainable parameters, with a pre-defined binary mask applied to it such that only the elements that correspond to an edge are optimized.

\subsection{Network Architecture}
The overall architecture of the proposed framework is illustrated in Fig. \ref{model network framework architecture}.
The number of stages $N$ is a hyper-parameter that is situational for model performance/complexity trade-off. In practice we find $N=2$ yields the best performance.
The modules each follow a encode--aggregate--decode paradigm. Each transitional pose prediction module has 3 graph convolutional layers, where the last one (the decoding layer) is specifically modified in order to produce only the desired transitional poses---See Eq. \ref{pose-wise graph convolution} for a detailed explanation. Each snippet-to-motion prediction module has multiple graph convolutional blocks, where the number of blocks and the dimension of node features are both hyper-parameters. The experiments show that the best performance is achieved when we employ 3 blocks for GCNs and use 128-dimensional node hidden features. We will analyze all of these hyper-parameters in ablation study.

\subsection{Training}
To train our model, we use the Mean Per Joint Position Error (MPJPE) proposed by \cite{ionescu2013human3}. 
Let $\hat{\tens{X}}_n$ be the $n$-th output sample generated by the model and $\tens{X}_n$ be the corresponding ground truth. Then the loss function for a total of $N$ training samples is defined as:
\begin{equation}
	\mathcal{L} = \frac{1}{NTJ}\sum_{n=1}^N \sum_{t=1}^T \sum_{j=1}^J \lVert (\hat{\tens{X}}_{t,j,:})_n - (\tens{X}_{t,j,:})_n \rVert_2.
\end{equation}
Besides MPJPE, we use another loss function to minimize the difference of the velocity, which is defined as:
\begin{equation}
	\mathcal{L}_\text{v} = \frac{1}{N(T-1)J}\sum_{n=1}^N \sum_{t=1}^{T-1} \sum_{j=1}^J \lVert (\hat{\tens{V}}_{t,j,:})_n - (\tens{V}_{t,j,:})_n \rVert_2,
\end{equation}
where
$\tens{V}_{t,j,:} = \tens{X}_{t+1,j,:} - \tens{X}_{t,j,:}.$
The overall training loss is:
\begin{equation}
    \mathcal{L}' = \mathcal{L} + \mathcal{L}_\text{v}.
\end{equation}
This loss function aims to minimize the $\ell_2$-norm between both the position and the velocity of the output and the ground truth.

    \begin{table}[t]
    \centering
    \resizebox{0.99\columnwidth}{!}{
    \begin{tabular}{p{2.6cm}|cccccc} \hline
    scenarios  &\multicolumn{6}{c}{Average}  \\ \hline
    milliseconds &80 &160 &320 &400 &560 &1000 \\ \hline
    LTD \cite{mao2019learning} &11.32 	&19.82 	&37.50 	&46.57 	&62.44 	&96.76 \\
    PGBIG \cite{ma2022progressively} &9.34 	&17.16 	&33.63 	&41.85 	&57.33 	&89.45 \\ \hline 
    Ours (non-shared) &11.00 &18.28 &34.27 &41.80 &57.29 &90.58 \\ 
    Ours (shared) &10.55 &17.92 & 33.87 &\textbf{\textcolor{blue}{41.24}} &\textbf{\textcolor{blue}{56.61}} &\textbf{\textcolor{blue}{88.01}}\\ \hline
    \end{tabular}}
    \caption{\small Average MPJPEs for prediction on CMU Mocap.}
    \label{cmu}
    \end{table}
    
    \begin{table}[t]
    \centering
    \resizebox{0.99\columnwidth}{!}{
    \begin{tabular}{p{2.6cm}|ccccc} \hline
    scenarios  &\multicolumn{5}{c}{Average}  \\ \hline
    milliseconds &200 &400 &600 &800 &1000 \\ \hline
    LTD \cite{mao2019learning} &36.81 &69.98 &92.42 &109.50 &120.96 \\
    PGBIG \cite{ma2022progressively} &35.36 &65.76 &86.79 &100.57 &109.12 \\ \hline 
    Ours (shared) &35.55 &66.32 &\textbf{\textcolor{blue}{85.71}} &\textbf{\textcolor{blue}{99.16}} &\textbf{\textcolor{blue}{106.24}} \\ \hline
    \end{tabular}}
    \caption{\small Average MPJPEs for prediction on 3DPW.}
    \label{3dpw}
    \end{table}

\section{Experiments}

\subsection{Datasets}
\textbf{Human3.6M} (H3.6M)
is a large-scale dataset \cite{ionescu2013human3} for human-related tasks. It involves 15 types of actions performed by 7 actors. Following the normality of previous work, we preprocess the data to remove global rotation and translation, discard redundant joints, and downsample the sequence to 25fps. Then we convert the data to 3D coordinates for the final training, validation and test sets. Each pose is characterized by 22 joints with their corresponding 3D coordinates. We use subject 11 and subject 5 for validation and testing respectively, and the remaining 5 subjects for training.

\noindent\textbf{CMU Motion Capture} (CMU Mocap)
consists of 5 general action categories. Following previous work, we use 8 specific types of actions, and perform data preprocessing such that each pose in the final data consists of 25 joints represented by 3D coordinates, and divide the data into training and test sets.

\noindent\textbf{3D Pose in the Wild} (3DPW)
covers activities captured from both indoor and outdoor scenes. We follow LTD \cite{mao2019learning} and use the official training, test and validation sets.
    
\subsection{Experimental Setup}

\textbf{Evaluation Metrics. }Following the evaluation paradigm in previous work, we report the experimental results in terms of the Mean Per Joint Position Error (MPJPE) in millimeters.

\noindent\textbf{Baselines. }
We compare the proposed framework with several methods that provide prediction results in terms of 3D error with released public codes, which include LTD \cite{mao2019learning}, MSR-GCN \cite{dang2021msr}, PGBIG \cite{ma2022progressively} and siMLPe \cite{guo2023back}.
For fair comparison we re-train them to obtain the results. Moreover, we adopt the same testing paradigm as previous state-of-the-art methods \cite{mao2019learning,li2020dynamic,martinez2017human}.

\noindent\textbf{Implementation Details. }
We implemented our model using Pytorch and Adam optimizer \cite{kingma2014adam} with learning rate 0.00001 with a 0.96 decay every 4 epochs. Our models were trained on NVIDIA RTX 3080 Ti GPU for 50 epochs, during which the batch size was 16 and the gradients were clipped to a maximum $\ell_2$-norm of 1.

    \begin{figure}[t]
    \centering
    \includegraphics[width=0.99\columnwidth]{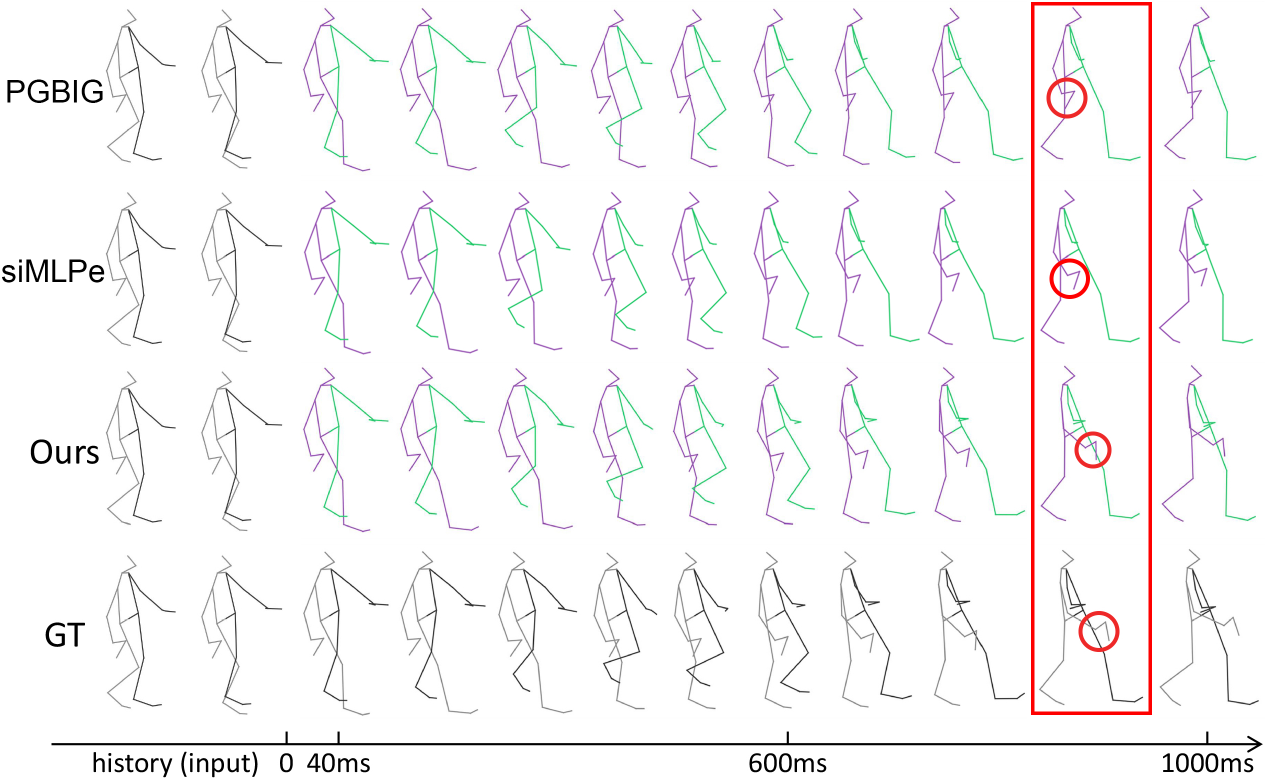} 
    \caption{\small Qualitative comparison on action “walking” in H3.6M. Compared to baselines, our method yields more accurate body movements, some of which are highlighted with red boxes and circles.}
    \label{viz}
    \end{figure}

\subsection{Comparison with State-of-the-Art Methods}
Consistently with the discussion above, in the evaluation of the proposed framework, we report the experimental results for motion prediction on H3.6M, CMU-Mocap and 3DPW. We are given 10 frames (400ms) to predict the future 25 frames. We also visualize the predicted poses for qualitative evaluation.

\textbf{Quantitative results on H3.6M.}
Table \ref{h36m_pred} shows the results on H3.6M. Following previous work, We compare the action-wise results of all 15 actions in 80ms, 400ms, 560ms, 1000ms respectively. It shows that our method in general. Compared with short-term prediction, long-term prediction is generally more challenging, because future poses are less deterministic and involve more uncertainty as time grows. Our method yields better results especially as time grows, which testifies to its power regarding long-term prediction.

\textbf{Qualitative results on H3.6M.}
For further comparison, we present the qualitative results in Figure \ref{viz}. We illustrate the future pose sequences predicted by ours and baseline methods including PGBIG and siMLPe, the most recent two state-of-the-art methods.
We provide an example case for walking. For the poses to be visually clear, we skip some poses every few frames.
Compared with baselines, our method generates more realistic and accurate poses. The poses generated by both PGBIG and siMLPe show imprecise body part movements, both of which therefore have accumulated large errors at the final pose. Ours gives better prediction results than those of others, especially with the increase of prediction time.

    \begin{figure*}[t]
    \centering
    \includegraphics[width=0.99\textwidth]{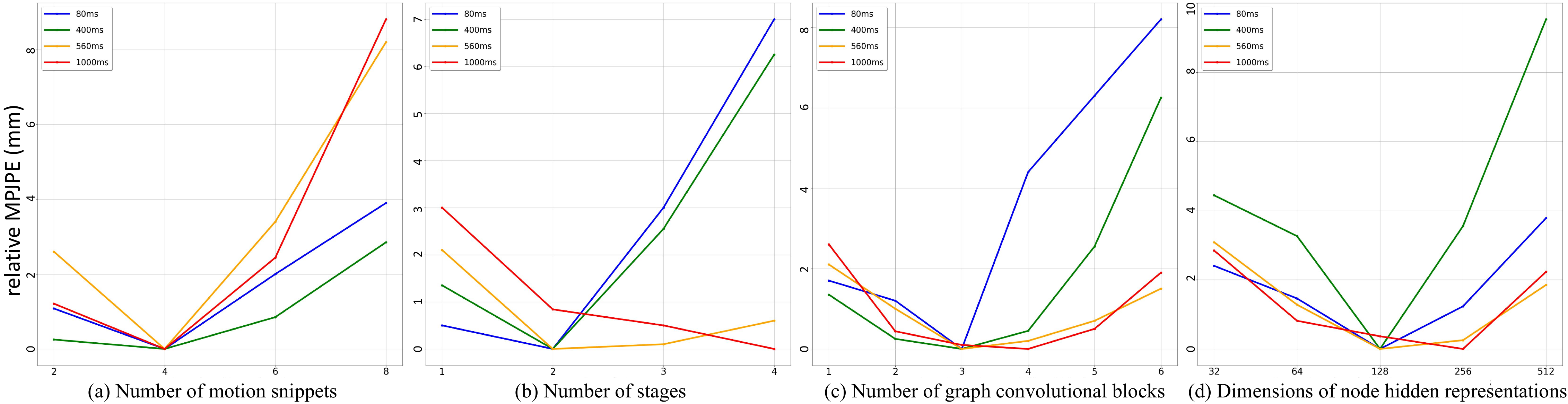}
    \caption{\small Ablation on multiple factors. (a) Effects of number of motion snippets. (b) Effects of number of stages. (c) Effects of number of GCN blocks. (d) Effects of dimension of node hidden features. The results are all reported in terms of relative MPJPE.}
    \label{ablation}
    \end{figure*}

    \begin{table*}[t]
    \centering
    \resizebox{0.85\textwidth}{!}{
    \begin{tabular}{p{4cm}|cc|p{4.5cm}|cc} \hline
    Reconstruction method &  Shared & Non-shared & Graph modeling &  Shared & Non-shared  \\ \hline
    linear interpolation (default) &  110.83 &\textbf{109.84}             & unified graph modeling (default) &  110.83 &\textbf{109.84} \\
    padding last frame & 111.27    &109.95                      & separate graph modeling & 111.60    & 110.01  \\
    padding first frame & 116.41    &113.36                     & spatial-only graph modeling & 117.56 & 116.33  \\ \hline
    \end{tabular}}
    \caption{\small Ablation on snippet reconstruction methods (left) and unified graph modeling (right). The MPJPE results at 1000ms are reported.}
    \label{ablation snippet reconstruction methods unified graph modeling}
    \end{table*}

\textbf{Quantitative results on CMU Mocap and 3DPW.}
Other than H3.6M dataset, we also evaluate our model on CMU Mocap and 3DPW, the results of which are shown in Table \ref{cmu} and \ref{3dpw}. Consistent with the evaluation setup reported in the papers of baseline methods, we report the averaged prediction errors of all activities at different timestamps. Compared to baselines, our method achieves the best performance on average on both datasets.

\subsection{Ablation Study}

To seek the determining factors of the advantage of our proposed snippet-to-motion progression framework, we present ablation studies on each essential component of our method.

\textbf{Number of transitional points.}
First we evaluate the performance of our methods in the cases of different numbers of transitional points or motion snippets. In Figure \ref{ablation} (a), lines of different colors correspond to different prediction timestamps, and we compare them in terms of relative MPJPE which means we subtract set the minimum value of each line as the basis and subtract them from all values. As shown in the figure, the best performance is obtained in the case of 4 motion snippets for each sample.

\textbf{Snippet reconstruction methods.}
Next we analyze different snippet reconstruction methods. As shown in Table \ref{ablation snippet reconstruction methods unified graph modeling}, the best method to reconstruct motion snippets is by applying linear interpolation between transitional poses, compared to padding the last or first pose of each snippet.

\textbf{Number of stages.}
To further evaluate the advantages of the proposed multi-stage framework, we implement different numbers of stages similar to PGBIG.
The results are presented in Figure \ref{ablation} (b) in terms of relative MPJPE same as other sub-figures. It shows that the case of 2 stages yields the best prediction result on average, albeit a little less optimal at 1000ms. The experiment shows that with more stages to implement the snippet-to-motion prediction framework, the model size and complexity grow rapidly, resulting in decrease in performance especially for short-term prediction.

\textbf{Unified graph modeling.}
The core backbone of the proposed framework is the graph convolutional network itself. The prediction depends on the graph convolution to extract features and aggregate information.
We first evaluate the advantages of our proposed unified graph modeling. As shown in Table \ref{ablation snippet reconstruction methods unified graph modeling}. Separate graph modeling refers to the kind adopted in methods such as PGBIG which constructs spatial and temporal graphs based on poses and trajectories separately. Spatial-only graph modeling refers to the kind as in LTD and MSR which aggregates features only depending on spatial adjacency. Both of the two methods will create indirect feature propagation paths and cause information loss, impairing their ability to capture long-range dependencies. The experiments verify the effectiveness of the unified graph modeling which shows better results than the other two.

\textbf{Number of graph convolutional blocks.}
Next we evaluate the structure of the graph network including the number of graph convolutional blocks, shown in Figure \ref{ablation} (c). We modify the number of blocks obtaining a series of differently structured networks, and show the test results for both long-term and short-term prediction. It shows that the prediction performance is the best when we employ 3 blocks to build each graph network.

\textbf{Dimension of node hidden representation.}
The key idea of graph convolution is that we want to learn hidden representations of nodes that depend on the structure of the graph in some higher-dimensional feature space.
Ablation on the dimension of node hidden representation in feature space is shown in Figure \ref{ablation} (d), where 128-dimensional features yield the best result.
In conclusion, our method achieves the best performance when we employ 4 motion snippets to split each sample, 2 stages for the framework, 3 graph convolutional blocks for each prediction module and 128-dimensional hidden representations to learn node features to implement our method.

\section{Conclusion}
In this paper we propose a novel snippet-to-motion prediction framework that breaks the motion prediction task into simpler sub-tasks.
The framework is inspired by the ubiquitous transitional patterns underlying diverse human motions, based on which we can split motion sequence into snippets characterized by their transitional poses.
We extend it to a multi-stage framework, with each stage tasked with transitional pose prediction, snippet reconstruction and snippet-to-motion prediction in sequential order. The first and last sub-tasks are modularized and implemented with graph convolutional networks, with guidance from intermediate supervision.
Considering that the framework is ultimately contingent on the functionality of the graph convolution network, we propose a novel graph representation to model motion sequence from a global perspective by dispensing with the space-time-separate graph construction adopted in current methods. Our unified graph modeling enables direct and efficient cross-spacetime feature aggregation and information propagation, which decidedly works to the advantage of our framework. Extensive experiments verify the effectiveness of the proposed framework.

\normalem
{\small
\bibliographystyle{ieee_fullname}
\bibliography{main}
}

\end{document}